\documentclass{aa}

\usepackage[varg]{txfonts}
\usepackage{hyperref} 
\usepackage{natbib}

\usepackage{algorithmic}
\usepackage{algorithm}
\usepackage{amssymb}
\usepackage{amsmath}
\usepackage{mathtools}
\usepackage{graphics}
\usepackage{graphicx,enumerate}

\DeclareGraphicsExtensions{.jpg,.png,.JPG, .bmp}
\usepackage{subfigure}
\usepackage{url}
\usepackage{bbm}

\usepackage{epsfig}
\usepackage{bm}
\usepackage[utf8]{inputenc}
\usepackage{color}
\usepackage{caption}
\usepackage[toc,page]{appendix}

\newcommand{\beqn}{\begin{eqnarray}}
\newcommand{\eeqn}{\end{eqnarray}}

\renewcommand{\ge}{\geqslant}
\renewcommand{\leq}{\leqslant}

\newcommand{\Id}{\mathrm{I}}

\DeclareMathOperator{\prox}{prox}

\DeclareMathOperator{\dom}{dom}
\DeclareMathOperator{\argmin}{argmin}
\DeclareMathOperator{\mad}{MAD}
\DeclareMathOperator{\diag}{\mathrm{diag}}
\DeclareMathOperator{\HT}{\mathrm{HardThresh}}
\DeclareMathOperator{\ST}{\mathrm{SoftThresh}}

\DeclareMathOperator*{\sinc}{sinc}
\DeclareMathOperator*{\s.t.}{s.t.}

\begin{document}

\title{Super-resolution method using sparse regularization for point-spread function recovery}

\author{F.M. Ngolè Mboula \thanks{fred-maurice.ngole-mboula@cea.fr}, J.-L. Starck, S. Ronayette, K. Okumura, J. Amiaux}

\institute{
Laboratoire AIM, UMR CEA-CNRS-Paris 7, Irfu, Service d'Astrophysique, CEA Saclay, F-91191 GIF-SUR-YVETTE Cedex, France}

\date{Received 11 June 2014;
Accepted 03 August 2014} 

\abstract{
In large-scale spatial surveys, such as the forthcoming 
ESA Euclid mission, images may be undersampled due to the optical sensors
sizes. Therefore, one may consider using a super-resolution (SR) method to recover
aliased frequencies, prior to further analysis. This is particularly relevant for point-source
images, which provide direct measurements of the instrument point-spread function (PSF). 
We introduce \textbf{SPRITE}, SParse Recovery of InsTrumental rEsponse, which is an SR algorithm using a sparse analysis prior. We show that such a prior
provides significant improvements over existing methods, especially on low SNR PSFs.
}
 

\titlerunning{Sparse PSF super-resolution}
\authorrunning{F.M. Ngolè Mboula et al.}

\maketitle

\section{Introduction}
The weak gravitational lensing is one of the most promising tools to probe the dark matter distribution in the universe. The idea is to infer, from a billion images of galaxies, the shape distortions due to dark matter gravitational lensing and then estimate the dark matter mass density along different lines of sight. 
The Euclid mission (launch planned for 2020) provides the data for such a purpose \citep{Eucl1}. Nevertheless, galaxy images are distorted due to the PSF. 
Therefore, it is critical to know this distortion accurately. It can be modeled in first approximation as a convolution of the desired image by the PSF of the telescope, which is typically space and time-varying. 
In practice, isolated stars provide PSF  measurements  at different locations in the field of view. 
Nevertheless, these stars images can be aliased as it is the case in Euclid, given the CCD sensor sizes.
On the other hand, the surveys are generally designed so that different images of the same stars are available and likely to be with different subpixel offsets on the sensor grid. Moreover, one may consider that nearby star images give to some extend the same local PSF. We, thus, can consider that different low-resolution versions of the same PSF are available in practice, so that one may apply an SR method to recover aliased frequencies.

This paper precisely tackles this problem. The SR is a widely studied topic in general image processing literature. Yet, some methods have been specifically proposed for astronomical data. For instance, there is the software IMCOM \citep{im_com} and PSFEx, which proposes an SR option. The IMCOM provides an oversampled output image from multiple undersampled input images, assuming that the PSF is fully specified. Since it does not deal with the PSF restoration itself, we use PSFEx as our main reference. The PSFEx performs SR by minimizing the sum of two terms. The first term is a weighted quadratic distance, relating the underlying PSF to each of the low resolution measurements. The second term consists of the square $l_2$ norm of the difference between the underlying PSF and a smooth first guess. This term is meant for regularization. In the proposed algorithm, we introduce a new regularization scheme based on the optimization variable sparsity in a suitable dictionary.

Section 2 presents the general principle of SR along with some state-of-the-art methods in astronomical domain. In Section 3, we present the proposed algorithm in details, which is followed with some numerical experiments in Section 4. We conclude by summarizing the main results and giving some perspectives. 

\section{Super-resolution overview}
\subsection{Notations}
We adopt the following notation conventions:
\begin{itemize}
\item we use bold low case letters for vectors;
\item we use bold capital case letters for matrices;
\item the vectors are treated as column vectors unless explicitly mentioned otherwise. 
\end{itemize}
The underlying high resolution (HR) image of size $d_1p_1\times d_2p_2$ is written in lexicographic order (for instance, lines after lines) as a vector of pixels values $\mathbf{x} = (x_1,...,x_q)^T$, where $q = d_1p_1d_2p_2$, and $d_1$ and $d_2$ are respectively the line and column downsampling factors. We consider $n$ LR observations. The vector of pixels values $\mathbf{y}_\mathbf{k} = (y_{k1},...,y_{kp})^T$ denotes the $k^{th}$ LR observation written in lexicographical order with $p = p_1p_2$ and $k=1...n$.

 \subsection{Observation model}
 \label{obs_mod_sec}
 
 We assume that $\mathbf{x}$ does not change during the acquisition of the $n$ LR images so that we have
	 
 \begin{equation}
 \mathbf{y_k} = \mathbf{D}\mathbf{B}_\mathbf{k}\mathbf{M}_\mathbf{k}\mathbf{x} + \mathbf{n}_\mathbf{k}, \; k=1...n.
\label{obs_mod_eq}
 \end{equation}
	
\begin{figure*}[ht!]
\begin{center}
\includegraphics[scale=0.55]{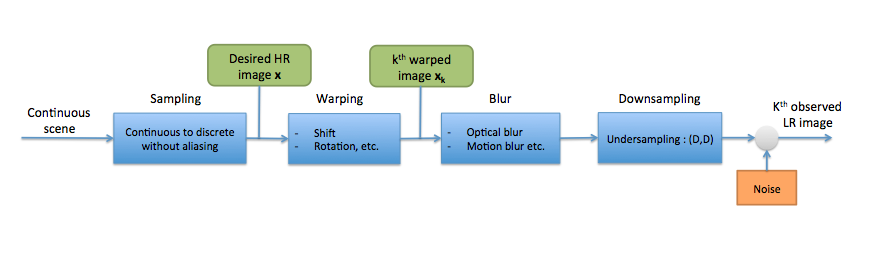}
\end{center}
\caption{General observation model. See Section \ref{obs_mod_sec} for a detailed description}
\label{obs_mod}
\end{figure*}

The variable $\mathbf{M}_\mathbf{k}$ is a warp matrix of size $q\times q$. It represents the motions of the observations relative to each other and those in general need to be estimated. The variable $\mathbf{B}_\mathbf{k}$ is a blur matrix of size $q\times q$. It accounts for different blurs (the atmosphere blur, which is particularly considerable for ground based telescopes, the system optics blur, the imaging system shaking etc.). The variable $\mathbf{D}$ is a matrix of size $p\times q$, which simply realizes a downsampling operation. Finally, $\mathbf{n}_\mathbf{k}$ is a noise vector of $q$ elements. This model is illustrated in Fig. \ref{obs_mod}.

\begin{figure*}[ht!]
\begin{center}
\includegraphics[scale=0.55]{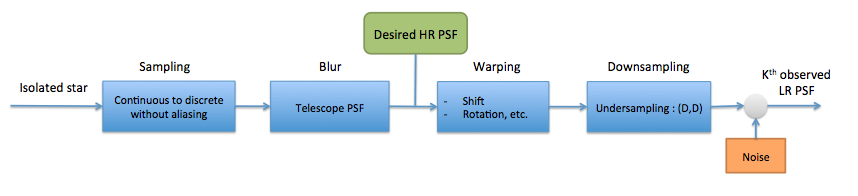}
\end{center}
\caption{Adapted observation model. This time we consider an isolated star as an input, and unlike in the general model, the output of the block "Blur" is the PSF.}
\label{obs_mod2}
\end{figure*}

In our case, we are interested in estimating the PSF, or in other terms, the telescope's contribution to the blur. We consider that the PSF varies slowly in the field so that the blocks "Warping" and "Blur" in Fig. \ref{obs_mod} may be swapped, for slow motions between observations. Therefore, the block diagram can be adapted as in Fig. \ref{obs_mod2}. This model still holds in presence of atmospheric blur (for ground-based telescopes) and jitter movements, if the LR images are extracted from the same exposure.  
	
In the general case, the model (\ref{obs_mod_eq})	 may simply be written as
\begin{equation}
\mathbf{y_k} = \mathbf{W}_\mathbf{k}\mathbf{x} + \mathbf{n}_\mathbf{k}, \; k=1...n,
\label{obs_mod_eq_1}
 \end{equation}
 where $\mathbf{W}_\mathbf{k}$ is a $p \times q$ matrix accounting for warping, blur, and downsampling. 
 
 \begin{figure*}[ht!]
\begin{center}
\includegraphics[scale=0.65]{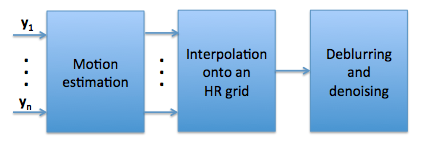}
\end{center}
\caption{SR general scheme. The deblurring does not apply to our case, since we want to precisely estimate the blur.}
\label{sr_scheme}
\end{figure*} 
 \subsection{SR techniques in astronomy}
 Generally, SR techniques involve three steps, which may be combined or performed separately. The registration step consists in evaluating the relative motions between different observations, so that their samples can be arranged on a common grid. It is critical that the precision of this registration should be smaller than the pixels dimensions over the target upsampling factors.   Since these relative motions are arbitrary, this grid is non-uniform. The next step would be to interpolate this grid in such a way to get a regularly sampled HR image. This image is blurry and noisy. Therefore, the final step is a restoration procedure. This scheme is summarized in Fig. \ref{sr_scheme}. In the next sections, we describe some SR techniques dedicated to astronomical images, but one may refer to \citet{kuo_sr} for more details on various SR frameworks in general image processing literature.	
 \subsubsection{Shift-and-add method}
 	\label{s_a_a}
 The most simple super-resolution method is certainly the shift-and-add method. It is performed in three steps. First, the images are upsampled to the target resolution. Then, they are shifted on a common grid and averaged. It has been used in astronomy for a long time, particularly for ground based telescopes. This method is simple, fast and is used for comparisons in the numerical experiments part. It has been shown that this method provides an optimal solution in the sense of the maximum likelihood (ML) with additive white Gaussian noise (WGN) and when only pure integer translation motions are considered with respect to the finer grid \citep{el1}.  The interpolation operator should be $\mathbf{D}^T$ (from Eq. \ref{obs_mod_eq}), which comes down to a simple zero-padding interpolation. It has been shown in the same work that the matrix $R = \sum_{k=1}^n \mathbf{M}_\mathbf{k}^T\mathbf{D}^T\mathbf{D}\mathbf{M}_\mathbf{k}$ is diagonal in this simple case. Thus, after registration and stacking of the interpolated images, each pixel value should be divided by the corresponding $R$ diagonal coefficient. 
  \subsubsection{PSFEx method}
  	\label{psfex_sec}
  The software PSFEx is an open source program, which has been used in many projects such as the Dark Energy Survey \citep{des} or CFHTLS\footnote{\mbox{\url{http://terapix.iap.fr/cplt/T0007/doc/T0007-doc.html}}} for PSF modeling. 
It takes a catalog of objects extracted from an astronomical image using SEXTRACTOR as input, which is also an open source tool for point sources (or stars) extraction. This catalog contains information about extracted point sources, such as SNR, luminosity, full width at half maximum (FWHM) , centroid coordinates, multiple flags related to saturation or blending, etc. Based on these measurements (performed in SEXTRACTOR) and some user provided parameters, PSFEx selects which sources are proper for PSF modeling. Afterwards, it constructs a PSF model, provides a fitting with an analytic function, and computes some of the PSF geometrical features. The PSF model construction may simply consist in optimally combining the input sources images for denoising, but it may also involve SR if these images are also undersampled. This SR functionality is our second reference for comparisons. These codes and the associated documentation may be found on the website \mbox{\url{http://www.astromatic.net/}}. 

The desired HR image is now a matrix $\mathbf{X}$ of size $dp\times dp$, where $d$ is the downsampling factor and the $k^{th}$ observation $\mathbf{Y}_\mathbf{k}$ is a matrix of size $p\times p$ with $k=1...n$. The coordinates of the centroid of the $k^{th}$ observation are denoted $(i_k,j_k)$. Assuming that the images are bandlimited, the samples of the LR images can be interpolated from the desired HR image, thanks to the Shannon sampling theorem \citep{shannon}. In theory, this interpolation should involve a 2D sinus cardinal (sinc) kernel with an infinite support, which is not convenient for practical implementation. One can instead use a support compact function, which approximates the sinus cardinal. Let $h(.,.)$ denote such a function. The estimate of the sample $(i,j)$ of the $k^{th}$ observation is given by
\begin{equation}
\hat{y}_{k,ij} = \sum_l\sum_m h\left[ l-d(i-i_k),m-d(j-j_k)\right]x_{ij}.
\label{intp_rel}
\end{equation}
Then, we can define the cost function 
\begin{equation}
J_1(\mathbf{X}) = \sum_{k=1}^n\sum_{i=1}^p\sum_{j=1}^p\frac{(y_{k,ij}-f_k \hat{y}_{k,ij})^2}{\sigma_k^2}
\label{psfex_cost}
\end{equation} 
where $f_k$ accounts for possible luminosity differences. The parameter $\sigma_k^2$ is related to the local background variance and any other uncertainty on the pixel value. These parameters need to be estimated.

The PSFEx uses a Lanczos4 interpolant, which is defined in 1D as 
\begin{equation}
h_{1D}(x) = \left\{
    \begin{array}{ll}
        1 & \mbox{if} \; x=0 \\
        \sinc(x)\sinc({x}/{4}) & \mbox{if}\; 0<|x|<4 \\
        0 & \mbox{else,}
    \end{array}
\right.
\end{equation}
so that $h(x,y) = h_{1D}(x)h_{1D}(y)$.

\begin{figure}[h]
\begin{center}
\includegraphics[scale=0.5]{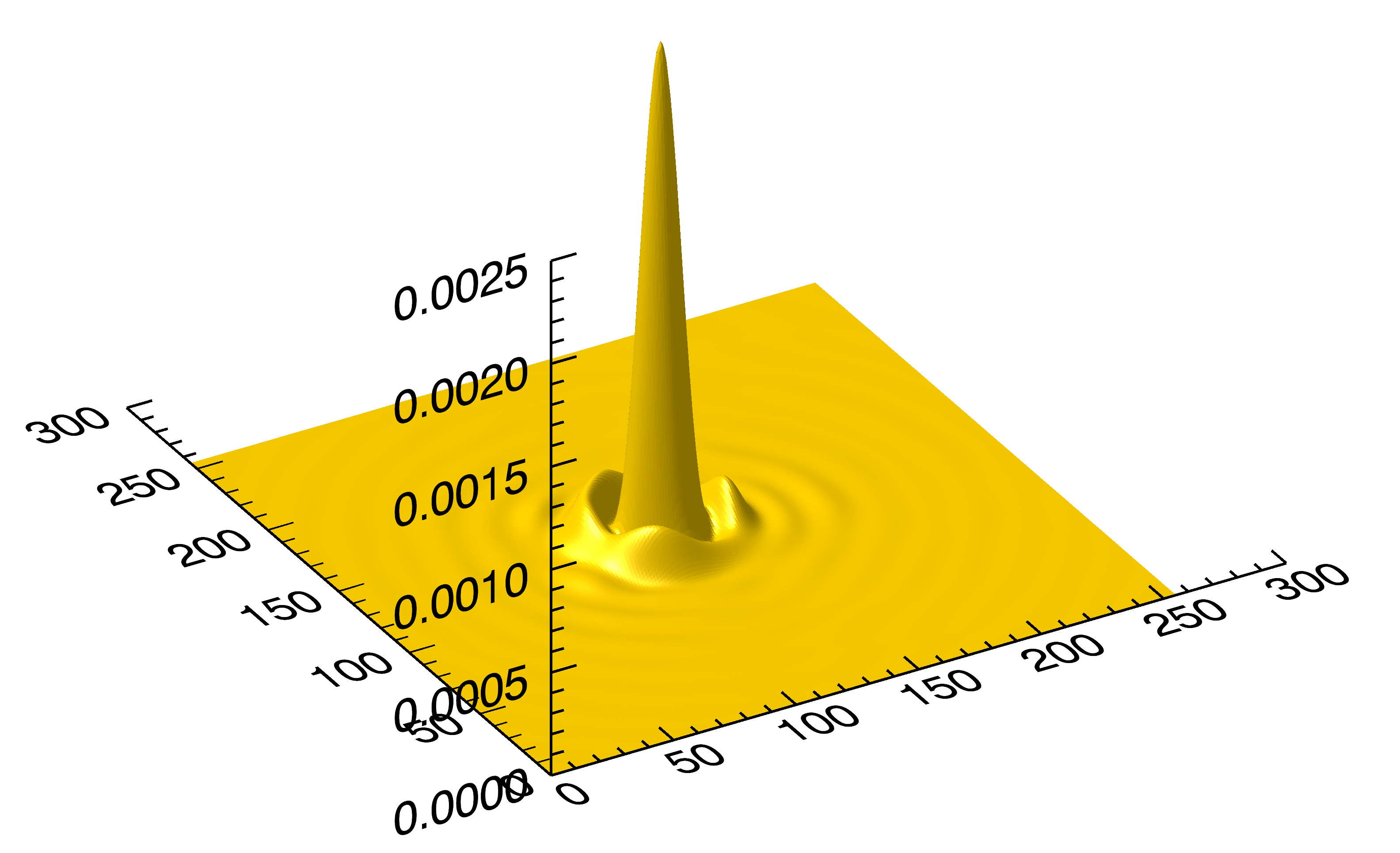} 
\end{center}
\caption{Simulated optical point-spread function.}
\label{PSF_view}
\end{figure}

Finally, PSFEx minimizes the cost function defined as
\begin{equation}
	\label{cost_func_reg}
J_2(\bm{\Delta}) = J_1(\bm{\Delta}+\mathbf{X}^{(0)}) + \lambda \|\bm{\Delta}\|_{2}^2,
\end{equation}
 where $\bm{\Delta} = \mathbf{X}-\mathbf{X}^{(0)}$ and $\mathbf{X}^{(0)}$ is a median image computed from the LR observations. This second term is meant for regularization purposes if the problem is ill-conditioned or undetermined ($n < d^2$).
One particularity of point-source images (see Fig. \ref{PSF_view}) is that one knows that the light comes from a single point, which is generally inferred to be the light blob's centroid. Thus, one only needs the images centroid coordinates to perform the registration, which is implicitly done in Eq. \ref{intp_rel}.
\section{Sparse penalty}
\subsection{Sparsity-based approaches}
Using a sparsity prior to regularize a linear inverse problem has proven very effective in a large variety of domains and, in particular, in astronomical data processing (see \cite{stk5} and the references therein). It has recently led to impressive results for 3D weak gravitational lensing mass map reconstruction \citep{glimpse}.
Let us consider the following general linear inverse problem: 
\begin{equation}
	\mathbf{y} = \mathbf{A}\mathbf{x} + \mathbf{n},
	\label{lin_inv}
	\end{equation}  
where $\mathbf{x} \in \mathbb{R}^p$ is the signal to recover, $\mathbf{y} \in \mathbb{R}^m$ are the noisy measurements, and $\mathbf{n}$ is an additive noise. The variable $\mathbf{A}$ is a matrix of size $m\times p$, which might be ill-conditioned, and in the general case $m$ can be smaller than $p$.
To regularize this problem, the signal $\mathbf{x}$ is assumed to be sparsely represented in an appropriate overcomplete dictionary $\bm{\Phi}$: $\mathbf{x} = \bm{\Phi}^T\bm{\alpha}$, where there is only a few non zero values in the vector $\bm{\alpha$}. We can define the $l_0$ pseudo-norm of $\bm{\alpha}$ as
	\begin{equation}
	\|\bm{\alpha}\|_0 = k,
	\end{equation} 
where k is the number of non-zero values in alpha.
Thus, one way to tackle the problem stated in Equation $\ref{lin_inv}$ under the sparsity assumption would be to solve the following
	\begin{equation}
	\min_{\bm{\alpha}} \|\bm{\alpha}\|_0 \s.t. \|\mathbf{y} - \mathbf{A}\bm{\Phi}^T\bm{\alpha}\|_2^2 < \epsilon,
		\label{l0_pb}
	\end{equation} 
where $\epsilon$ is related to the noise variance. 
In practice, for most problems, the signals of interest are not strictly sparse, but this assumption can be relaxed to compressible signals; when expressed in a suitable dictionary, the coefficients of a compressible signal exhibit a polynomial decrease of their sorted absolute values:
\begin{equation}
	|\alpha_{(i)}| \leq C i^{-\frac{1}{b}},\;\;1\leq i\leq q,\;b \in ]0,1]
\end{equation}
where $C$ is a constant. This tends to be verified for a well-chosen dictionary.

Besides, the problem \ref{l0_pb} is combinatorial and is untrackable in most practical applications. One instead uses the $l_1$ norm as a relaxation of the $l_0$ pseudo-norm and solves a convex optimization problem of the form: 	
\begin{equation}	
	\min_{\bm{\alpha}} \frac{1}{2}\|\mathbf{y} - \mathbf{A}\bm{\Phi}^T\bm{\alpha}\|_2^2 + \lambda\|\bm{\alpha}\|_1,
\label{bpdn_lag}
	\end{equation} 
where the parameter $\lambda$ balances the sparsity against the data fidelity.
This formulation is known as the augmented Lagrangian form of the basis pursuit denoising (BPDN) problem \citep{dnh4}. In solving this problem, one seeks a sparse way to synthesize the wanted signal from the atoms of the dictionary $\bm{\Phi}$. This prior is referred to as the synthesis prior in the sparse recovery literature. Another way of promoting sparsity is through the analysis prior, which consists of seeking a solution, which has a sparse representation in the transform domain, without imposing the signal to be written as a linear combination of some dictionary atoms. This is done by solving the following problem: 
    \begin{equation}	
	\min_{\mathbf{x}} \frac{1}{2}\|\mathbf{y} - \mathbf{A}\mathbf{x}\|_2^2 + \lambda\|\bm{\Phi}\mathbf{x}\|_1,
\label{analys_prior}
	\end{equation} 
	where the matrix $\bm{\Phi}$ transforms $\mathbf{x}$ in the new representation domain.
	The problems \ref{bpdn_lag} and \ref{analys_prior} are not equivalent when the matrix $\bm{\Phi}$ is not unitary. Therefore, a choice has to be made between the two, which is discussed in the next section.

\subsection{Method}
\label{method}
Now, let consider Eq. \ref{intp_rel}. It can be rewritten as
\begin{equation}
\hat{\mathbf{y}}_k = \mathbf{D}\mathbf{H}_k\mathbf{x},
\label{intp_rel_mat}
\end{equation}
where $\mathbf{x}$ and $\hat{\mathbf{y}}_k$ are, respectively, the desired matrix and the $k^{th}$ observation estimate written this time as column vectors in lexicographic order (we used the lines order); besides, $\mathbf{H}_k$ is a Toeplitz matrix \citep{toep} of size $d^2p^2\times d^2p^2$, which contains the values of the kernel $h(.,.)$ appearing in Eq. \ref{intp_rel} and $\mathbf{D}$ is a decimation matrix of size $p^2\times d^2p^2$ with $d$ being the downsampling factor. We are assumming that the images are squared for convenience but they might be non-squared as well.
In the same way, we can redefine the objective function of Equation \ref{psfex_cost} as
 \begin{equation}
J_1(\mathbf{x}) = \frac{1}{2}\sum_{k=1}^n{\|\mathbf{y}_\mathbf{k}-f_k\mathbf{D}\mathbf{H}_k\mathbf{x}\|_2^2}/{\sigma_k^2},
\label{psfex_cost_mat}
\end{equation} 
where $\mathbf{y}_\mathbf{k}$ is the $k^{th}$ observation rewritten consistently with $\mathbf{x}$ and $\hat{\mathbf{y}}_k$. The function $J_1(\mathbf{x})$ is nothing but the log-likelihood associated with the observation model in the case of an uncorrelated Gaussian noise that is stationary for each observation up to a scalar factor. It can be written in an even more compact way as
\begin{equation}
J_1(\mathbf{x}) = \frac{1}{2}\|\bm{\Sigma}^{-1}\mathbf{y}-\bm{\Sigma}^{-1}\mathbf{F}\mathbf{W}\mathbf{x}\|_2^2,
\label{psfex_cost_mat_2}
\end{equation} 
where $\mathbf{W}$ is obtained by concatening vertically the matrices $\mathbf{D}\mathbf{H}_k$, $\mathbf{F}$ is a diagonal matrix constructed by repeating the coefficients $f_k$ $p^2$ times for $k=1...n$, and $\bm{\Sigma}$ is constructed the same way using the coefficients $\sigma_k$.
Therefore, we can simply write
\begin{equation}
J_1(\mathbf{x}) = \frac{1}{2} \|\mathbf{z}-\mathbf{M}\mathbf{x}\|_2^2,
\label{psfex_cost_mat_3}
\end{equation} 
where $\mathbf{M}$ is a matrix of size $np^2\times d^2p^2$. 
The SPRITE method constrains the minimization of this objective function using an analysis prior. Instead of using a single lagrangian multiplier as in Eq. \ref{analys_prior}, the analysis coefficients has individual weights $\kappa\lambda_i$, where $i$ is the coefficient index in the transform domain. This leads to the following formulation of the problem:
\begin{equation}
\underset{\bm{\Delta}}\min J_1(\bm{\Delta}+\mathbf{x}^{(0)}) +\kappa\|\bm{\lambda}\odot\bm{\Phi}\bm{\Delta}\|_1,
\label{sr_sp1}
\end{equation}
where $\bm{\Delta} = \mathbf{x}-\mathbf{x}^{(0)}$ is defined as in \ref{cost_func_reg}, $\mathbf{x}^{(0)}$ is a first guess, and $\bm{\lambda}$ is now a vector of the same size as $\bm{\Phi}\bm{\Delta}$, $\odot$ denoting the pointwise product. 

Additionally, the PSF or equivalently the telescope optical impulse response is by definition a positive valued function \citep{thomps}. Therefore, we want the reconstruction $\mathbf{x}$ to have positive entries. This additional constraint is integrated as follows:
\begin{equation}
\underset{\bm{\Delta}}\min J_1(\bm{\Delta}+\mathbf{x}^{(0)}) +\kappa\|\bm{\lambda}\odot\bm{\Phi}\bm{\Delta}\|_1 \; s.t. \bm{\Delta} \ge -\mathbf{x}^{(0)},
\label{sr_sp1p}
\end{equation}
where $\ge$ is a pointwise inequality, or equivalently,
\begin{equation}
\underset{\bm{\Delta}}\min J_1(\bm{\Delta}+\mathbf{x}^{(0)}) +\kappa\|\bm{\lambda}\odot\bm{\Phi}\bm{\Delta}\|_1 + \mathbbm{1}_{\mathcal{S}_{\mathbf{x}^{(0)}}}(\bm{\Delta}),
\label{sr_sp1p2}
\end{equation}
where $\mathcal{S}_{\mathbf{x}^{(0)}}$ is the set of vectors $\mathbf{t} \in \mathbb{R}^{d^2p^2}$ satisfying the pointwise inequality $\mathbf{t} \ge -\mathbf{x}^{(0)}$ and $\mathbbm{1}_{\mathcal{S}_{\mathbf{x}^{(0)}}}$ is its indicator function (see Appendix \ref{prox_app}). The impact of this constraint is emphasized in Appendix \ref{pos_cons}.

As we show in the Section \ref{param_set}, the choice of the parameters $\bm{\lambda}$ and $\kappa$ relies on the noise expected on the analysis coefficients of the solution estimate. The choice of a vector regularization parameter rather than a single scalar is precisely motivated by the fact that this noise might be non-stationary.
 
As stated before, the problem \ref{sr_sp1} is not equivalent to its synthesis version, if the dictionary $\bm{\Phi}$ is redundant. The synthesis prior is expected to be efficient if the desired solution can be accurately written as a sparse linear combination of the chosen dictionary atoms, which we cannot assume to be true for every PSF profiles. In contrast, the analysis prior appears to be more flexible. Moreover, in the cases where the problem would be ill-conditioned or underdetermined, the analysis prior would definitely be more suitable since it involves far less variables. 

A similar $l_1$ penalty has already been applied for SR. An example may be found in \citet{Iyam}, where the cost function is minimized using variants of the alternating direction method of multipliers (ADMM). Moreover, advantages of such approaches over quadratic regularizers have been shown in many related problems in image and signal processing.

The use of the $l_1$ norm as a relaxation for an $l_0$ penalty has a well-known drawback, which is that it tends to bias the solution. Indeed with a $l_1$ norm penalty, the problem resolution involves soft thresholding operations, which affect both weak and strong entries unlike a hard thresholding, which would only affect the weak and therefore unwanted entries. Formal definitions of soft and hard thresholding are given in Appendix \ref{prox_app}.

This is particularly unsuitable for scientific data analysis. The reweighting $l_1$ minimization proposed in  \cite{Cand} is one way to tackle this issue, while staying in the proof of convergence sets. Indeed, it consists of solving a succession of $l_1$ minimization problems of the form  
\begin{equation}
\underset{\bm{\Delta}}\min J_1(\bm{\Delta}+\mathbf{x}^{(0)}) +\kappa\|\mathbf{w}^{(k)}\odot\bm{\lambda}\odot\bm{\Phi}\bm{\Delta}\|_1 + \mathbbm{1}_{\mathcal{S}_{\mathbf{x}^{(0)}}}(\bm{\Delta}),
\label{sr_sp_rw}
\end{equation}
where $\mathbf{w}^{(k)}$ is a weighting vector for the transform coefficients at the $k^{th}$ minimization. Each entry of $\mathbf{w}^{(k)}$ is calculated as a decreasing function of the corresponding transform coefficient magnitude in the $(k-1)^{th}$ minimization. This way, the strong transform coefficients are less penalized than the weaker ones in the new minimization. One may refer to Appendix \ref{rw} for quantitative study of the reweighting effect.

\subsection{Algorithm}
\label{algorithm}
In the proposed method, the reweighting scheme is performed according to \citet{Cand}: 
\begin{enumerate}
  \item Set $k = 0$, for each entry in the weighting vector $\mathbf{w}^{(k)}$, set $w_j^{(k)} = 1$.
  \item Solve the problem \ref{sr_sp_rw} yielding a solution $\bm{\Delta}^{(k)}$.
  \item Compute $\bm{\alpha}^{(k)} = \bm{\Phi}\bm{\Delta}^{(k)}$.
  \item Update the weight vector according to $w_j^{(k+1)} = \frac{1}{1+|\alpha_j^{(k)}|/3\sigma_j}$, where $\sigma_j$ is the noise standard deviation expected at the $j^{th}$ transform coefficient, see section \ref{param_set}. 
  \item Terminated on convergence or when reaching the maximum number of iterations; otherwise, go to step 2.
\end{enumerate}

The step 2 resolution is detailed in Algorithm \ref{sprite_ana}. We use the generalized forward-backward splitting introduced in \citet{rag}. It requires the computation of proximity operators associated with the regularization functions in \ref{sr_sp_rw}. One may refer to Appendix \ref{prox_app} for an introduction to proximal calculus.

\begin{algorithm*}[!htb]
\caption{Weighted analysis-based $\bm{\Delta}^{(k)}$ recovery}
\label{sprite_ana}
\begin{algorithmic}[1]
\REQUIRE $\quad$ \\
A first guess estimate of the super-resolved image $\mathbf{x}^{(0)}$. \\
A weight vector $\mathbf{w}^{(k)}$. \\ 
Sparsity constraint parameter $\kappa$. \\
A dictionary $\bm{\Phi}$. \\
Auxiliary variables $\mathbf{z}_{10}, \mathbf{z}_{20} \in \mathbb{R}^{d^2p^2}$.\\
$\omega_1,\omega_2 \in ]0,1[ \; s.t. \; \omega_1+\omega_2=1, \lambda>0$ (see Section \ref{algo_param}).	
\bigskip

\STATE Initialize $\mathbf{d}_0 = \omega_1 \mathbf{z}_{10} + \omega_2 \mathbf{z}_{20}$.
\FOR{$n=0$ to $N_{\max}-1$} 
 \STATE $\mathbf{z}_{1n+1} = \mathbf{z}_{1n} + \lambda (\prox_{\frac{\mu}{\omega_1}\kappa\|\diag{(\mathbf{w}_k\odot\bm{\lambda})}\bm{\Phi}\|_1}{(2\mathbf{d}_n-\mathbf{z}_{1n} - \mu\nabla J_1(\mathbf{d}_n+\mathbf{x}^{(0)})}-\mathbf{d}_n)$
 \STATE $\mathbf{z}_{2n+1} = \mathbf{z}_{2n} + \lambda (\prox_{\frac{\mu}{\omega_2}\mathbbm{1}_{\mathcal{S}_{\mathbf{x}^{(0)}}}(\bm{\Delta})}{(2\mathbf{d}_n-\mathbf{z}_{2n} - \mu\nabla J_1(\mathbf{d}_n+\mathbf{x}^{(0)})}-\mathbf{d}_n)$
 \STATE $\mathbf{d}_{n+1} = \omega_1 \mathbf{z}_{1n+1} + \omega_2 \mathbf{z}_{2n+1}$	
 \ENDFOR
 \STATE {\bf Return:} $\bm{\Delta}^{(k)} = \mathbf{d}_{N_{\max}}$.
\end{algorithmic}
\end{algorithm*} 

Since the dictionary $\bm{\Phi}$ is redundant, we do not have a closed-form expression for the proximity operator. Yet, it can be calculated as 
\begin{equation}
\left\lbrace \begin{matrix}\prox_{\frac{\mu}{\omega_1}\kappa\|\diag{(\mathbf{w}^{(k)}\odot\bm{\lambda})}\bm{\Phi}\|_1}{(\mathbf{x})} = \mathbf{x} - \bm{\Phi}^T\hat{\mathbf{u}} \\
\hat{\mathbf{u}} = \argmin_{|u_j| < \frac{\mu}{\omega_1}\kappa w_j^{(k)} \lambda_j} \frac{1}{2}\|\mathbf{x}-\bm{\Phi}^T\mathbf{u}\|_2^2 
\end{matrix}\right.\,
\label{prox_anl}
\end{equation}
where $\hat{\mathbf{u}}$ can be estimated using a forward-backward algorithm \citep{Cmb1} as follows:
\begin{enumerate}
  \item Set $p = 0$, initialize $\mathbf{u_0}=0$.
  \item $\tilde{\mathbf{u}}_{p+1} = \mathbf{u}_p +\mu_{\prox}\bm{\Phi}\left(\mathbf{x}-\bm{\Phi}^T\mathbf{u_{p}}\right)$.
  \item $\mathbf{u}_{p+1} = \tilde{\mathbf{u}}_{p+1} - \ST_{\frac{\mu}{\omega_1}\kappa\mathbf{w}^{(k)}\odot\bm{\lambda}}\tilde{\mathbf{u}}_{p+1}$.
  \item Terminate on convergence or when reaching the maximum number of iterations, otherwise go to step 2.
\end{enumerate}
The thresholding operator $\ST$ is defined in Appendix \ref{prox_app}. The operator $\prox_{\frac{\mu}{\omega_2}\mathbbm{1}_{\mathcal{S}_{\mathbf{x}^{(0)}}}}$ is simply the orthogonal projector onto the set $\mathcal{S}_{\mathbf{x}^{(0)}}$ defined in the previous section; it is given explicitly Appendix \ref{prox_app}.

A full description of SPRITE is provided in Algorithm \ref{sprite_ana_2}. 

\begin{algorithm*}[!htb]
\caption{SPRITE: weigthed analysis-based super-resolution}
\label{sprite_ana_2}
\begin{algorithmic}[1]
\REQUIRE $\quad$ \\
Sparsity constraint parameter $\kappa$. \\
A dictionary $\bm{\Phi}$.\\
An upsampling factor.
\bigskip
\STATE Estimate the data fidelity parameters (see Section \ref{dat_fid_param}).
\STATE Calculate a first guess $\mathbf{x}^{(0)}$.
\STATE Initialize $\mathbf{w}^{(k)} = 1$.
\STATE Calculate a step size $\mu$.
\FOR{$k=0$ to $K_{\max}-1$} 
 \STATE $\bm{\Delta}^{(k)}=\underset{\bm{\Delta}}\argmin J_1(\bm{\Delta}+\mathbf{x}^{(0)}) +\kappa\|\mathbf{w}^{(k)}\odot\bm{\lambda}\odot\bm{\Phi}\bm{\Delta}\|_1 + \mathbbm{1}_{\mathcal{S}_{\mathbf{x}^{(0)}}}(\bm{\Delta})$ (see Algorithm \ref{sprite_ana})
 \STATE $\bm{\alpha}^{(k)} = \bm{\Phi}\bm{\Delta}^{(k)}$
 \STATE $w_j^{(k+1)} = \frac{1}{1+|\alpha_j^{(k)}|/3\sigma_j}$
\ENDFOR
\STATE {\bf Return:} $\hat{\mathbf{x}} = \bm{\Delta}^{(K_{\max}-1)}+\mathbf{x}^{(0)}$.
\end{algorithmic}
\end{algorithm*} 
  
  \subsection{Parameter estimation}
  	The data fidelity term in the problem \ref{sr_sp_rw} is defined as  
\begin{equation}
\left\lbrace \begin{matrix}J_1(\mathbf{x}) = \frac{1}{2}\sum_{k=1}^n\sum_{i=1}^p\sum_{j=1}^p\frac{(y_{k,ij}-f_k \hat{y}_{k,ij})^2}{\sigma_k^2} \\
	\hat{y}_{k,ij} = \sum_l\sum_m h\left[ l-d(i-i_k),m-d(j-j_k)\right]x_{ij}
	\end{matrix}\right.,
\label{sprite_cost}
\end{equation} 
where $\mathbf{x} = (x_{ij})_{1 \leq i,j \leq dp}$ is the desired image and $h(.,.)$ is a 2D Lanczos kernel.
As stated in Section \ref{method}, it can be written as
\begin{equation}
J_1(\mathbf{x}) = \frac{1}{2}\|\mathbf{z}-\mathbf{M}\mathbf{x}\|_2^2,
\label{psfex_cost_mat_3bis}
\end{equation} 
if we write $\mathbf{x}$ in lexicographic order as a vector.
The following parameters are required:
\begin{itemize}
	\item for the data fidelity term parameters the noise standard deviation in the LR images $\sigma_k$, the photometric flux $f_k$, and the shift parameters $(i_k,j_k)$;
	\item a first guess $\mathbf{x}^{(0)}$ (see problem \ref{sr_sp_rw});	
	\item the sparsity constraint parameters $\kappa$, $\bm{\lambda} = (\lambda_i)_i$ and the dictionary $\bm{\Phi}$;
	\item algorithmic parameter such as the gradient step size $\mu$, the relaxation parameter $\lambda$ in Algorithm \ref{sprite_ana}, and the gradient step size $\mu_{\prox}$ in Equation \ref{prox_anl} resolution. 	
\end{itemize}

  \subsubsection{Data fidelity parameters}
  	\label{dat_fid_param}
  	\paragraph{Noise standard deviations}
At the first step of the algorithm, the noise standard deviations in the low resolution images can be robustly estimated using the median absolute deviation (MAD) estimator \citep{stk5}.
	\paragraph{Subpixel shifts}
The subpixel shifts between the images are estimated based on the low resolution images centroids positions. Those are calculated on the low resolution images after a hard thresholding operation. For the image $\mathbf{x}_i$, the threshold is chosen as 
\begin{equation}
k = \min(4\sigma_i,\left(\frac{\max(|\mathbf{x}_i|)}{\sigma_i}-1\right)\sigma_i).
\end{equation}
In this way, we only keep pixels with a high SNR for the centroid estimation. We then estimate the centroid positions using the iteratively weighted algorithm introduced in \cite{Baker}. The thresholding operation undoubtedly biases the estimated centroid position, but the resulting estimated shifts are expected to be unbiased, up to the finite sampling and noise effects. 

	\paragraph{Photometric flux}	
The flux parameters are calculated by integrating the low resolution images on a fixed circular aperture centered on their centroids estimates. At Euclid resolution (see Section \ref{num_exp}), we obtained quite accurate flux estimates in simulations using a radius of 3 pixels for the aperture.
These parameters define the matrix $M$ in Eq. \ref{psfex_cost_mat_3bis}.

\bigskip
All these parameters are automatically calculated without requiring any user input. 

\subsubsection{First guess computation}
\label{first_guess_comp}
As one can see in the Algorithm \ref{sprite_ana_2}, the final image is computed as 
\begin{equation}
\hat{\mathbf{x}} = \bm{\Delta}^{(K_{\max}-1)}+\mathbf{x}^{(0)}.
\end{equation}
This implies that the noise and any artifact in $\mathbf{x}^{(0)}$ which does not have a sparse decomposition in $\bm{\Phi}$ will be present in the final solution. Therefore, one has to be careful at this step. To do so, we compute a noisy first guess $\mathbf{x}_n^{(0)}$ using a shift-and-add, as presented in Section \ref{s_a_a}. Then we apply a wavelet denoising to $\mathbf{x}_n^{(0)}$. In other terms, we transform $\mathbf{x}_n^{(0)}$ in a "sparsifying" wavelet dictionary $\mathbf{W}$. We threshold each wavelet scale in such a way to keep only the coefficients above the noise level expected in the scale. Finally, we apply a reconstruction operator, which is a dictionary $\widehat{\mathbf{W}}$ verifying $\widehat{\mathbf{W}}\mathbf{W} = \Id$ to the thresholded coefficients (see \citet{stk5}). We note $\bm{\beta} = (\beta_i)_i$, a vector made of the denoising thresholds for each wavelet scale. We set $\beta_i$ at $5\sigma_i$, where $\sigma_i$ is the noise standard deviation in the $i^{th}$ wavelet scale. The first guess is finally computed as 
\begin{equation}
\label{mod_den}
\mathbf{x}^{(0)} =  \widehat{\mathbf{W}}\HT_{\bm{\beta}}\mathbf{W}\mathbf{x}_n^{(0)},
\end{equation}
which robustly removes the noise without breaking important features.
One can refer to Appendix \ref{first_guess_wav_noise} for $(\sigma_i)_i$ estimation. 
\subsubsection{The choice of dictionary and regularization parameter}
\label{param_set}
\paragraph{Regularization parameter}
The regularization parameter $\kappa$ can be set, according to a desired level of significance. Indeed, it can be seen that the transform domain vector $\hat{\mathbf{u}}$ is constrained into weighted $l^\infty$-ball of radius $\mu\kappa$ in Equation \ref{prox_anl} and can be interpreted as the non-significant part of the wanted signal current estimate. To set this radius according to the expected level of noise for each transform coefficient, we propagate the noise on the data vector $\mathbf{z}$ from Equation \ref{psfex_cost_mat_3bis} through $\mu\bm{\Phi}\mathbf{M}^T\mathbf{M}$ and estimate its standard deviation at each transform coefficient, which sets the parameters $\lambda_j$. In practice, this can be done in two ways. We can either run a Monte-Carlo simulation of the noise in $\mathbf{z}$ and take the empirical variance of the sets of realizations of each transform coefficient. On the other hand, if $\bm{\Phi}$ is a wavelet dictionary and if the noise is expected to be stationary in each wavelet scale, then we only need to compute a single standard deviation per scale. This can be done by estimating the noise in each scale of the wavelet transform of the gradient at each iteration (up to the factor $\mu$) using a MAD, for instance. Indeed, the residual $\mathbf{z}-\mathbf{M}(\mathbf{d}_n+\mathbf{x}^{(0)})$ tend to be consistent with the noise in $\mathbf{z}$, so that it can be used as a noise realization. With a stationary noise in each input image, the two approaches give very close estimates of the noise standard deviation and the second one is far less demanding in terms of complexity. As a result, coefficients below $\kappa\lambda_j$ are considered as part of the noise and one only needs to set the global parameter $\kappa$  to tune the sparsity constraint according to the noise level. 
\paragraph{Dictionary} 
 The choice of the dictionary impacts the performance of the algorithm. We considered two transforms: a biorthogonal undecimated wavelet transform with a 7/9 filter bank and the second generation starlet transform \citep{starlet}. These two transforms are generic and not specifically tuned to a given PSF profile. 

\subsubsection{Algorithmic parameters}
\label{algo_param}

\paragraph{Gradient steps sizes}
The gradient step size $\mu$ in Algorithm \ref{sprite_ana} needs to be chosen just in $\left]0,2/\rho(\mathbf{M}^T\mathbf{M})\right[$, where $\rho(.)$ denotes the spectral radius of a square matrix. In the same way, $\mu_{\prox}$ needs to be chosen in $\left]0,2/\rho(\bm{\Phi}\bm{\Phi}^T)\right[$.

\paragraph{Relaxation parameter}
The parameter $\lambda$ in Algorithm \ref{sprite_ana} needs to be chosen in $\left]0,\min\left(\frac{3}{2},\frac{1+2/{\rho(\mathbf{M}^T\mathbf{M})\mu}}{2}\right)\right[$ \citep{rag}. This parameter tunes the updating speed of the auxiliary variables in Algorithm \ref{sprite_ana}. In practice, we use $\mu = 1/\rho(\mathbf{M}^T\mathbf{M})$ and $\lambda = 1.4$.

\subsubsection{User parameters}
\label{user_param} 
It is important to mention that the user only has to set the parameter $\kappa$ and the dictionary with the other parameters being automatically estimated. In all our experiments, we took $\kappa = 4$, which is quite convenient, if we assume Gaussian noise in the data. The dictionary choice will be emphasized in the next section.

\section{Numerical experiments}
\label{num_exp}
This section presents the data used, the numerical experiments realized as a mean to compare three SR techniques (shift-and-add, PSFEx method, and our method) and the results. 
\subsection{Data set}		
The PSFs provided are optical PSFs computed using a fast Fourier transform of the exit pupil. They are not a system PSF, so they do not include a jitter or detector response. A set of PSFs covering the whole field of view is provided. They are monochromatic PSFs at 800nm and are derived from tolerance analysis. They account for manufacturing and alignments errors and thermal stability of the telescope. Manufacturing and alignment errors are partially compensated by a best focus optimization, while thermal stability effects are simulated by a small displacement of the optics that are not compensated on a short-time scale.
The optical model used is dated from 2011 and is prior to the current reference model (provided by Astrium, which has been awarded the payload module contract in 2013). In particular, the 2011 model does not contain the latest definition of the pupil mask. The pupil, however, includes central obscuration and a three-vane spider. This is the model that has been used for the science feasibility studies that led to the acceptance of the Euclid mission.
\subsection{Simulation}
		
In the Euclid mission, the actual sampling frequency is about $0.688$ times the Nyquist frequency that we define as  twice the telescope spatial cut-off frequency \citep{Crop1}. Therefore we target an upsampling factor of $2$, which gives a sufficient bandpass to recover the high frequencies.
The PSF is typically space-varying, and this is particularly true for wide field of view instruments as Euclid telescope \citep{Eucl1}. Thus, the data set contains simulated PSF measurements on a regular $18\times 18$ grid on the field of view. 
The original PSF models are downsampled to twice Euclid resolution. For each PSF, four randomly shifted "copies" are generated and downsampled to Euclid resolution (see Fig. \ref{HR_PSF} and Fig. \ref{lr_PSF} below). These LR images are of size $84\times 84$. Different levels of WGN are added. We define the signal level as its empirical variance that is calculated in a $50 \times 50$ patch centered on the HR PSF main lobe. For each algorithm, we used these four images to reconstruct a PSF which has twice their resolutions in lines and columns. 
\subsection{Quality criterion}
For an image $\mathbf{X} = (x_{ij})_{i,j}$, the weighted central moments are defined as 
\begin{equation}
\mu_{p,q}(\mathbf{X}) = \sum_i\sum_j(i-i_c)^p(j-j_c)^p f_{ij}x_{ij} 
\end{equation}
with $(p,q) \in \mathbb{N}^2$, $(i_c,j_c)$ are the weighted image centroid coordinates, and $\mathbf{F} = (f_{ij})_{i,j}$ is an appropriate weighting function (typically a Gaussian function).
The ellipticity parameters are then defined as follows:

\begin{gather}
e_1(\mathbf{X}) = \frac{\mu_{2,0}(\mathbf{X})-\mu_{0,2}(\mathbf{X})}{\mu_{2,0}(\mathbf{X})+\mu_{0,2}(\mathbf{X})}  \\
e_2(\mathbf{X}) = \frac{2\mu_{1,1}(\mathbf{X})}{\mu_{2,0}(\mathbf{X})+\mu_{0,2}(\mathbf{X})}.
\label{ellipticity}
\end{gather}

The vector $\bm{\epsilon} = \left[e_1,e_2\right]$ is an important tool, since if measured on a large set of galaxies, it can be statistically related to the dark matter induced geometrical distortions and finally its mass density. Furthermore, this ellipticity parameters are magnitude invariant and approximately shift invariant. The error on ellipticity is therefore an interesting criteria for quality assessment. 
Thus, we used the mean absolute error for each ellipticity parameter, 
\begin{equation}
\mathcal{E}_j = \frac{1}{n}\sum_{i=1}^n |e_j(\mathbf{X}_\mathbf{i})-e_j(\mathbf{\hat{X}}_\mathbf{i})|, \; j=1,2
\label{ell_err}
\end{equation}
and the associated empirical standard deviations.

Moreover, the PSF size is also an important characteristic of the PSF kernel. For example, it has been shown in \citet{paulo} that the PSF size largely contributes to the systematic error in weak gravitational lensing surveys. Therefore, we use it in quality assessment by computing the mean absolute error on the full width at half maximum (FWHM). The FWHM is estimated by fitting a modified Lorentzian function on the PSF images. We used routines from a publicly available library\footnote{\mbox{\url{http://www.astro.washington.edu/docs/idl/htmlhelp/slibrary21.html}}}.

\subsection{Results and discussion}
\begin{figure}[h]
\begin{center}
\includegraphics[scale=0.45]{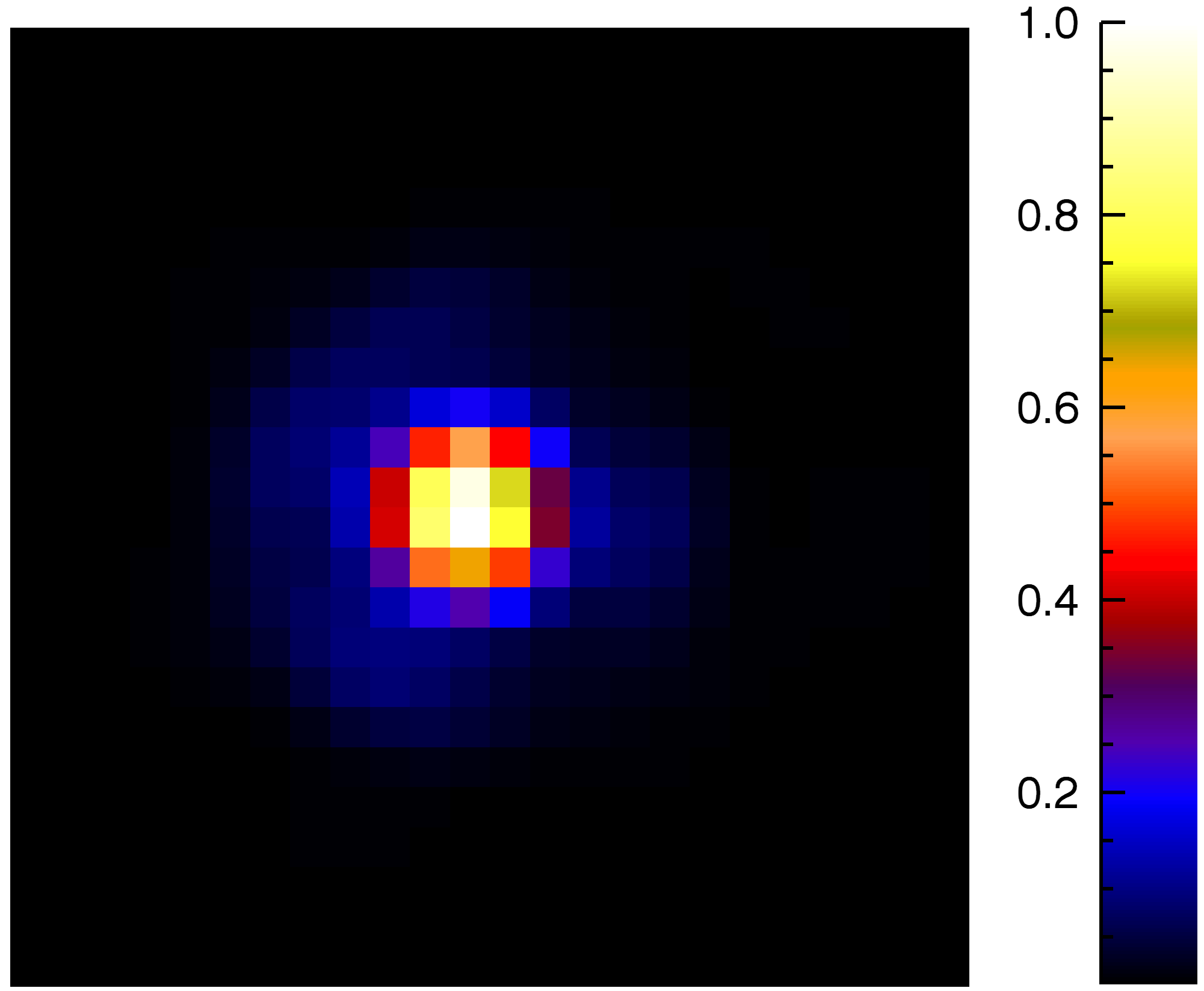}
\end{center}
\caption{Critically sampled PSF.}
\label{HR_PSF}
\end{figure}

\begin{figure}[h]
\begin{center}
\includegraphics[scale=0.45]{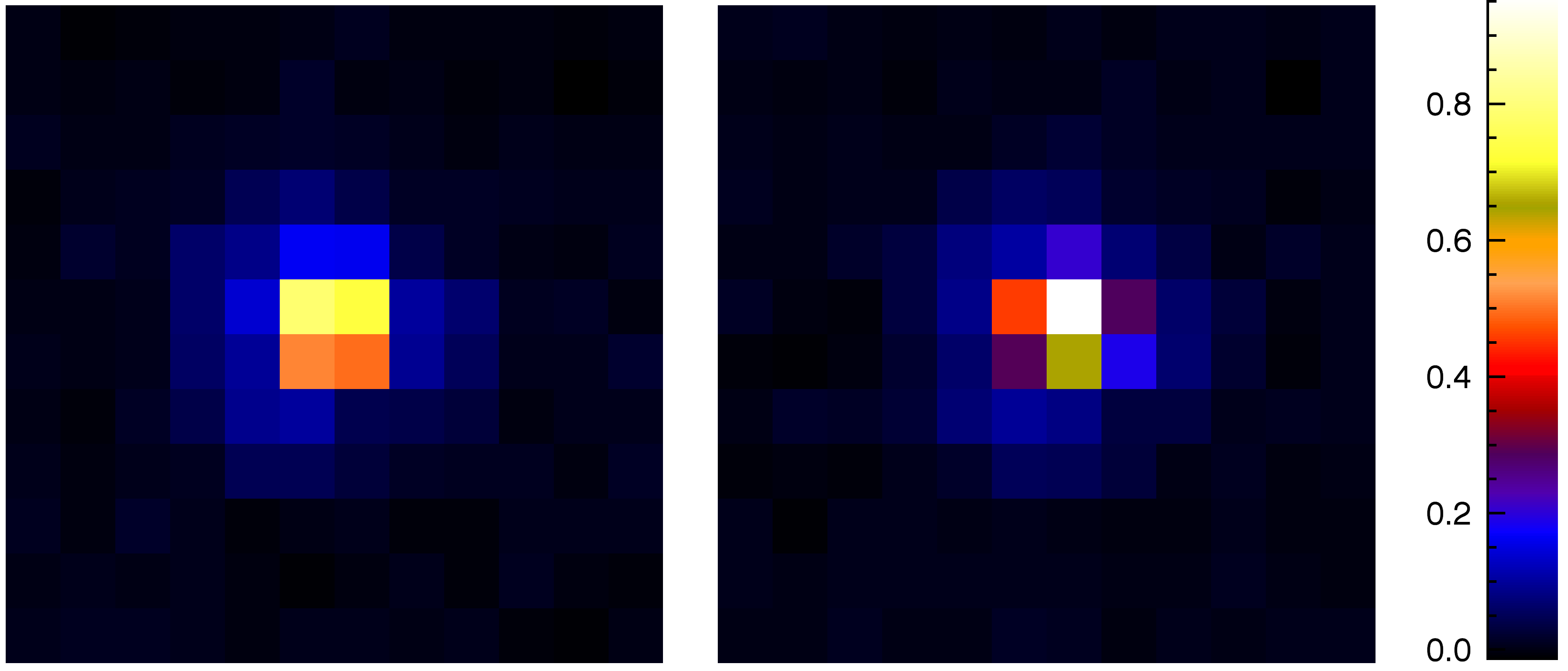}
\end{center}
\caption{PSF sampled at Euclid resolution with different offsets and noise.}
\label{lr_PSF}
\end{figure}

The Figures \ref{HR_PSF} and \ref{lr_PSF} show a simulated PSF that is sampled at almost the Nyquist rate and the two LR shifted and noisy PSF derive, with SNR of around 30dB.

\begin{figure*}[ht!]
\begin{center}
\begin{tabular}{ccc}
\subfigcapskip = 5pt
\subfigure[Super-resolved PSF]{\includegraphics[width = 0.80\textwidth]{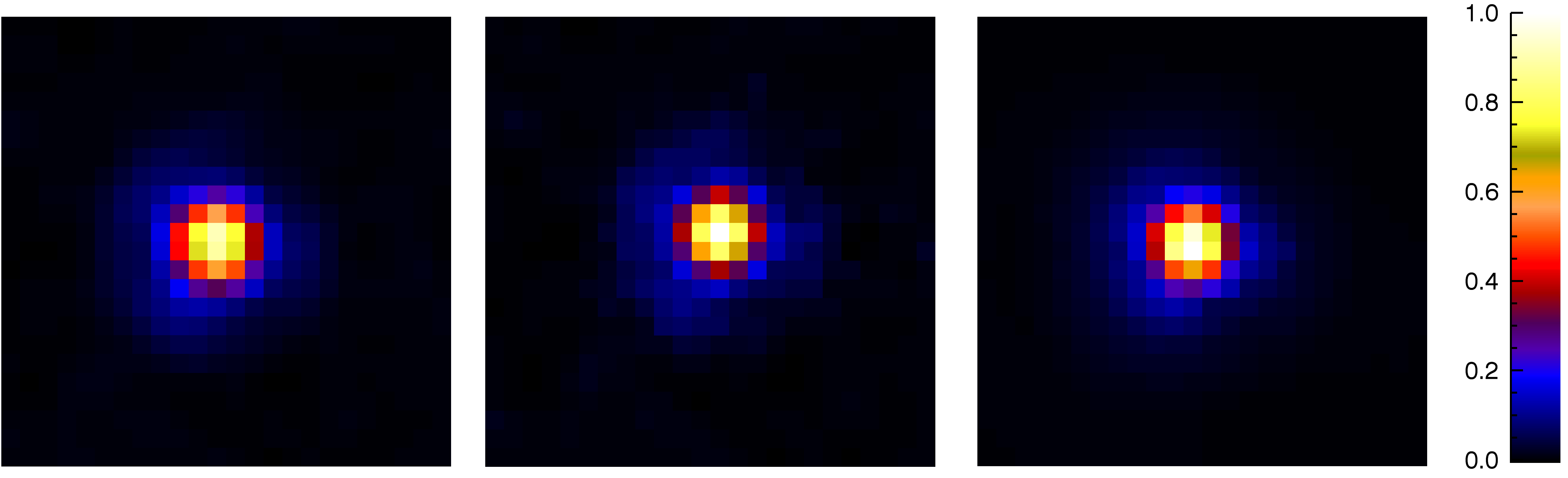}} \\
\label{first_guess_den_b}	
\subfigcapskip = 5pt
\subfigure[Absolute Error Maps]{\includegraphics[width = 0.80\textwidth]{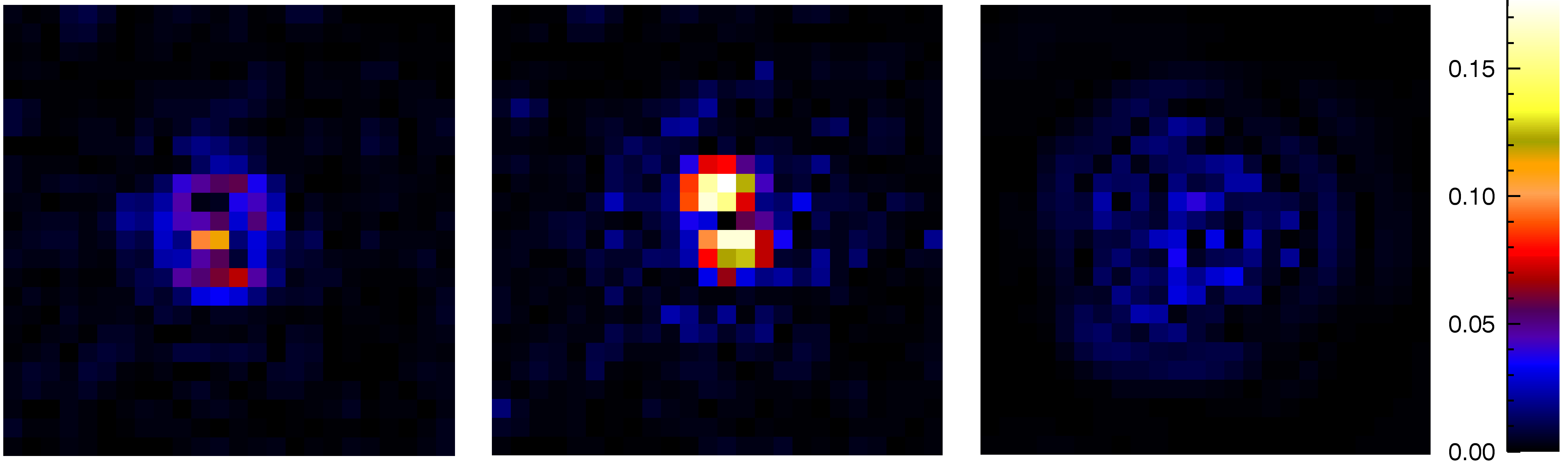}} 
\end{tabular}
\end{center}
\caption{PSF reconstruction and error map at 30dB for three methods: from the left to the right, shift-and-add, PSFEx, and SPRITE. The error image standard deviation is at least 30\% smaller with SPRITE.}
\label{psf_rec}
\end{figure*}

The Figure \ref{psf_rec} shows an example of super-resolved PSF at 30dB of SNR, from four LR images and the corresponding error maps that are defined as the absolute value of the difference between the original high resolution noise free PSF and the PSF reconstructions for each algorithm. This error map standard deviation is at least $30\%$ lower with SPRITE.
  
\begin{figure*}[ht!]
\begin{center}
\begin{tabular}{ccc}
\subfigcapskip = 5pt
\subfigure[Mean absolute error on the first ellipticity parameter.]{\includegraphics[width=7cm]{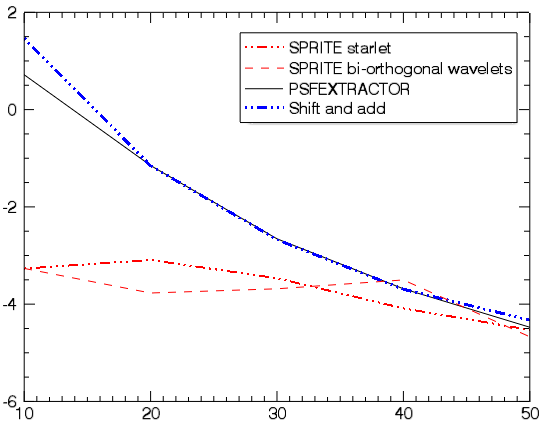}} &
\label{first_guess_den_b}	
\subfigcapskip = 5pt
\subfigure[Standard deviation of the absolute error on the first ellipticity parameter.]{\includegraphics[width=7cm]{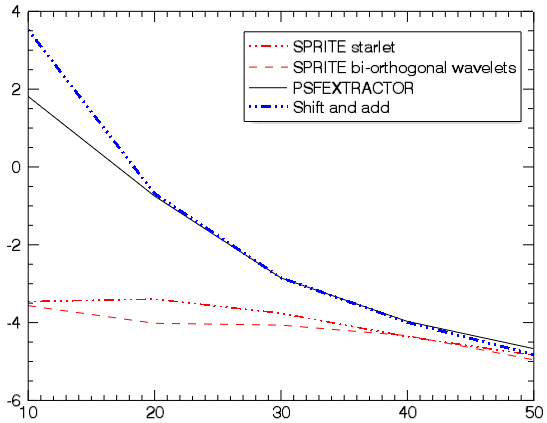}} \\
\subfigcapskip = 5pt
\subfigure[Mean absolute error on the second ellipticity parameter.]{\includegraphics[width=7cm]{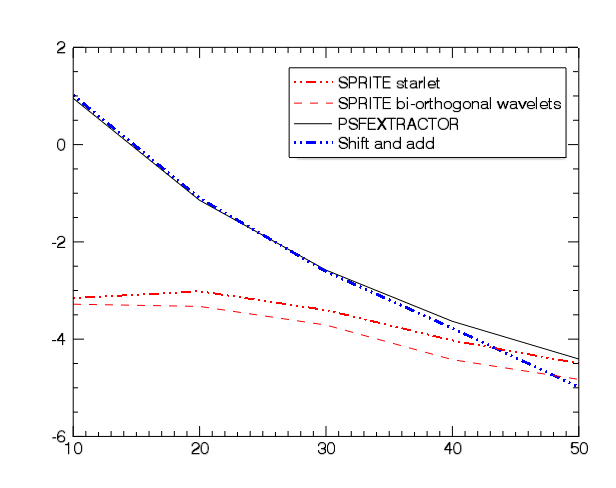}} &
\label{first_guess_den_b}	
\subfigcapskip = 5pt
\subfigure[Standard deviation of the absolute error on the second ellipticity parameter.]{\includegraphics[width=7cm]{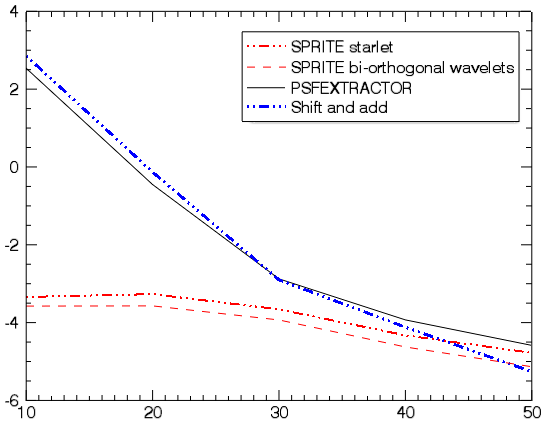}} 
\end{tabular}
\end{center}
\caption{Errors in log on ellipticity parameters versus the SNR. For SNR=10dB, SPRITE achieves around 6dB less than others methods, which corresponds to a factor of $e^6$ on a linear scale.}

\label{num_res_1}
\end{figure*}

\begin{figure}[h]
\begin{center}
\includegraphics[scale=0.4]{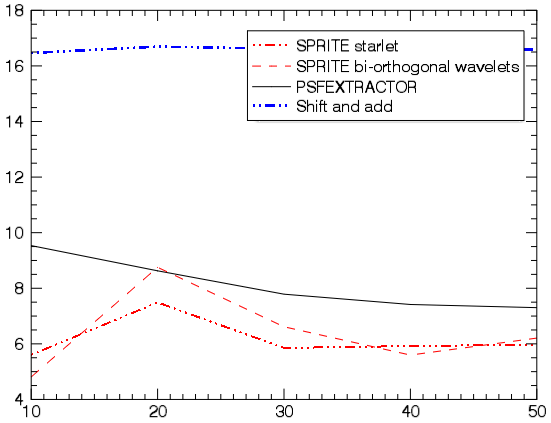}
\end{center}
\caption{ Mean absolute error on the full width at half maximum (FWHM) in percent. SPRITE achieves on average an error of $6\%$ on the FWHM which is $2\%$ less than PSFEX in average.}
\label{fwhm}
\end{figure}

Figure \ref{num_res_1} shows smaller errors and errors dispersions are achieved with SPRITE algorithm, especially at low SNR.  One can note that the dispersion is slightly smaller with biorthogonal undecimated wavelet. The error on the FWHM given in percent on Fig. \ref{fwhm} is smaller with SPRITE. In practice, there is more variability in the PSF (wavelength and spatial dependency, time variations...) so that the real problem will be more underdetermined. Thanks to the multiple exposures on the one hand, and that the spatial variations of the PSF are expected to be slow on the other hand, the real problem could actually be very well constrained. Moreover, these results suggest that even better results could be achieved by using more adapted dictionaries,  built either from PSF model or through a dictionary learning algorithm \citep{simon}. 

\section{Complexity and performances}
The simulations were run on a typical desktop computer. Let us suppose that we have $n$ LR images of sizes $p_1 \times p_2$ and that we choose an upsampling factor $d$ in lines and columns. As stated before, we took $p_1  = p_2 = 84$, $n = 4$, and $d = 2$ in our numerical experiments. Under this setting, it takes roughly $60s$ and $1$GB of physical memory to compute a super-resolved PSF. More generally, the computational complexity of the algorithm is in $O(n p_1 p_2 d^2\log{}(p_1 p_2 d^2))$, which is related to the implementation of the matrices $\mathbf{M}$ from Eq. \ref{psfex_cost_mat_3} and $\mathbf{M}^T$ using FFT.

\section{Software}
Following the philosophy of reproducible research \citep{reprod_res}, the algorithm introduced in this paper and the data used are available at \mbox{\url{http://www.cosmostat.org/sprite.html}}. We used the following calls for the SPRITE executable:
\begin{itemize}
\item \textbf{run\textunderscore sprite -t 2 -s 4 -r 2 -F -N data\textunderscore file output\textunderscore file output\textunderscore directory} for the second genration Starlet transform;
\item \textbf{run\textunderscore sprite -t 24 -s 4 -r 2 -F -N data\textunderscore file output\textunderscore file output\textunderscore directory} for the undecimated biorthogonal wavelet transform.
\end{itemize}
The options "-s" and "-r" set the parameter $\kappa$ (see Section \ref{param_set}) and the upsampling factor for both lines and columns respectively. The options "-F" and "-N" indicate that the photometric flux and the noise might have different levels in the LR images and need to be estimated.

\section{Conclusion}
We introduced SPRITE, which is a super-resolution algorithm based on sparse regularization. We show that adding a sparse penalty in the recovery leads to far better accuracy in terms of ellipticity error, especially at low SNR. 

Quantitatively, we achieved
\begin{itemize}
\item a $30\%$ lower error on the reconstruction itself at $30$dB of SNR; 
\item around 6dB less than other methods on the shape parameters, which corresponds to a factor of $e^6$ on a linear scale, at $10$dB of SNR;
\item $6\%$ of error on the FWHM in average, which $2\%$ less than PSFEX.
\end{itemize}

\bigskip

However this algorithm does not handle the PSF spatial variations. Thus one natural extension of this work would be to simultaneously perform super-resolution and dimensionality reduction assuming only one LR version of each PSF, as it is the case in practice strictly speaking, but by using a large PSF set. The PSF wavelength dependency would also be an interesting aspect to investigate. 

\appendix
\section{First guess noise}
We keep the same notations as in Section \ref{first_guess_comp}. We note $\mathbf{x}_{wi}$ as the $i^{th}$ scale of $\mathbf{x}_n^{(0)}$ wavelet transform in $\mathbf{W}$. As the noise in $\mathbf{x}_n^{(0)}$ is correlated, the estimation of  $\sigma_i$ is not straightforward, so we proceed as follows:
\begin{enumerate}
\item $\sigma_i^{0} = 1.4826\mad(\mathbf{x}_{wi})$ 
\item $\hat{\mathbf{x}}_{wi} = \underset{\mathbf{x}}\argmin \frac{1}{2}\|\mathbf{x}-\mathbf{x}_{wi}\|_2^2 + k\sigma_i^{0}\|\mathbf{x}\|_1$
\item $\sigma_i = 1.4826\mad(\widehat{\mathbf{x}}_{wi}-\mathbf{x}_{wi})$.
\end{enumerate}
The factor $1.4826$ comes from the assumption that the noise is approximately Gaussian. Finally, the factor $k$ must be sufficiently high so that all the noise will remain in the residual $\widehat{\mathbf{x}}_{wi}-\mathbf{x}_{wi}$. We took $k = 5$. The minimization in step 2 usually takes up to five iterations to converge. For the finer scales, $\sigma_i^{0}$ is quite close to $\sigma_i$. But for coarser scales, $\sigma_i^{0}$ is significantly overestimated (it might be more than 10 times greater than $\sigma_i$). We implicitly assumed that the noise in the wavelet scales is stationary, which is reasonable apart from the edges effects due to the wavelet transform. 

\label{first_guess_wav_noise}

\section{Proximal calculus}
\label{prox_app}
Let $\mathcal{H}$ be a finite-dimensional Hilbert space (typically a real vector space) equipped with the inner product $\langle . , . \rangle$ and associated with the norm $\|.\|$. A real-valued function $\mathcal{F}$ defined on $\mathcal{H}$ is 
\begin{itemize}
\item proper if its domain, as defined by $\dom{\mathcal{F}} = \{\mathbf{x} \in \mathcal{H} / \mathcal{F}(x) < +\infty\}$, is non-empty; 
\item lower semicontinuous (LSC) if $\lim \inf_{\mathbf{x}\rightarrow \mathbf{x}_0}\mathcal{F}(\mathbf{x}) \geq \mathcal{F}(\mathbf{x}_0)$.
\end{itemize}	
We define $\Gamma_0(\mathcal{H})$ as the class of all proper LSC convex real-valued function defined on $\mathcal{H}$.
	
	Moreau (1962) introduced the notion of proximity operator as a generalization of a convex projection operator. Let $\mathcal{F} \in \Gamma_0(\mathcal{H})$. Then the function $\mathbf{y} \rightarrow \frac{1}{2}\|\bm{\alpha}-\mathbf{y}\|^2+\mathcal{F}(\mathbf{y})$ achieves its minimum at a unique point denoted by $\prox_{\mathcal{F}}(\bm{\alpha})$, $(\forall \bm{\alpha} \in \mathcal{H})$. The operator $\prox_{\mathcal{F}}$ is the proximity operator of $\mathcal{F}$. The indicator function of a closed convex subset $\mathcal{C}$ of $\mathcal{H}$ is the function defined on $\mathcal{H}$ by 
\begin{equation}
	\mathbbm{1}_{\mathcal{C}}(\mathbf{x}) = \left\lbrace \begin{matrix} 0, \; \text{if} \; \mathbf{x} \in \mathcal{C} \\
	+\infty, \; \text{otherwise}. 
\end{matrix}\right.\
\end{equation}	
It is clear from the definitions that the proximity operator of $\mathbbm{1}_{\mathcal{C}}$ is the orthogonal projector onto $\mathcal{C}$. Thus, for $\mathcal{C} = \mathcal{S}_{\mathbf{x}^{(0)}}$, which is defined in Section \ref{method}, and for $\mathbf{x} \in \mathbb{R}^{d^2p^2}$, we have,
\begin{equation}
\prox_{\mathbbm{1}_{\mathcal{S}_{\mathbf{x}^{(0)}}}}(\mathbf{x}) = (\max(x_i,-x^{(0)}_i))_{1\leq i \leq d^2p^2}.
\end{equation}

	Now we suppose that $\mathcal{H} = \mathbb{R}^p$, and we want to solve 
	\begin{equation}
	\underset{\bm{\alpha} \in \mathbb{R}^p }\min \mathcal{F}_1(\bm{\alpha}) + \mathcal{F}_2(\bm{\alpha})
	\label{GenConv}
	\end{equation}
	where $\mathcal{F}_1, \mathcal{F}_2 \in \Gamma_0(\mathcal{\mathbb{R}}^p)$. Many problems in signal and image processing may be formulated this way, where $\mathcal{F}_1$ would be the data attachment function and $\mathcal{F}_2$ would constrain this solution based on prior knowledges. It has been shown \citep{Cmb2} that if $\mathcal{F}_1$ is differentiable with a $\beta$-Lipschitz continuous gradient, then the problem (\ref{GenConv}) admits at least one solution and that its solutions, for $\gamma >0 $, verify the fixed point equation, 
	\begin{equation}
	\mathbf{x} = \prox\nolimits_{\gamma\mathcal{F}_2}(\mathbf{x} - \gamma\nabla\mathcal{F}_1(\mathbf{x})).
	\end{equation}	
	This suggests the following iterative scheme,
	\begin{equation}
	\mathbf{x}_{n+1} = \prox_{\gamma_n\mathcal{F}_2}(\mathbf{x}_n - \gamma_n\nabla\mathcal{F}_1(\mathbf{x}_n)),
	\end{equation}
	 for appropriate values of the parameter $\gamma_n$. This type of scheme is known as forward-backward (FB) algorithm : a forward gradient step using $\mathcal{F}_1$ and a backward step involving  $\mathcal{F}_2$ through its proximity operator.
	 Some examples of FB algorithms that have been shown to converge to a solution of (\ref{GenConv}) can be found in \citet{Cmb1}.
	 A popular example of proximity operator is the one associated with $\mathcal{F} = \lambda\|.\|_{1}$ with $\lambda \in \mathbb{R}$: 
\begin{equation}
\begin{multlined}
\prox_{\lambda\|.\|_1}(\bm{\alpha}) = \ST_\lambda(\bm{\alpha}) = \\
\shoveleft[1cm] \left(\left(1 - \frac{\lambda}{|\alpha_i|}\right)_+\alpha_i\right)_{1 \leq i \leq p},
\end{multlined}
\end{equation}
where $p$ is the dimension of $\mathcal{H}$, $\alpha_i$ the components of alpha in the basis associated with $\|.\|_{l1}$ and $(.)_+ = \max(.,0)$.
One may refer to \citet{Cmb1} for proximity operator properties and examples. For $\bm{\lambda} = (\lambda_i)_{1 \leq i \leq p}$, the proximity operator associated with the weighted $l_1$ norm $\|\diag(\bm{\lambda})(.)\|_{1}$ is given by 
\begin{equation}
\begin{multlined}
\prox_{\|\diag(\bm{\lambda})(.)\|_{1}}(\bm{\alpha}) = \ST_{\bm{\lambda}}(\bm{\alpha}) = \\
\shoveleft[1cm] \left(\left(1 - \frac{\lambda_i}{|\alpha_i|}\right)_+\alpha_i\right)_{1 \leq i \leq p}.
\end{multlined}
\end{equation}
The hard thresholding operator defined as
\begin{equation}
\begin{multlined}
\HT_{\bm{\lambda}}(\bm{\alpha}) = (\tilde{\alpha}_i)_{1 \leq i \leq p}, \\
\shoveleft[1cm]\text{with} \; \tilde{\alpha}_i = \left\{
    \begin{array}{ll}
        \alpha_i & \mbox{if} \; |\alpha_i|\geq \lambda_i \\
        0 & \mbox{else.}
    \end{array}
\right.
\end{multlined}
\end{equation} 
is often used instead in practice. See \citet{Blum} for more insight on the hard thresholding behavior in terms of convergence.

\section{Positivity constraint}
\label{pos_cons}
\begin{figure*}[ht!]
\begin{center}
\begin{tabular}{ccc}
\subfigcapskip = 5pt
\subfigure[Mean absolute error on the first ellipticity parameter.]{\includegraphics[width=7cm]{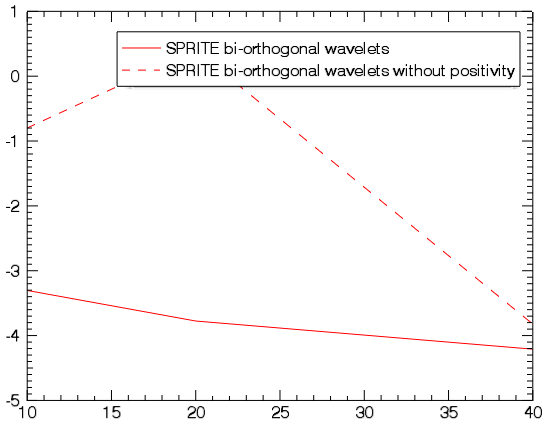}} &	
\subfigcapskip = 5pt
\subfigure[Standard deviation of the absolute error on the first ellipticity parameter.]{\includegraphics[width=7cm]{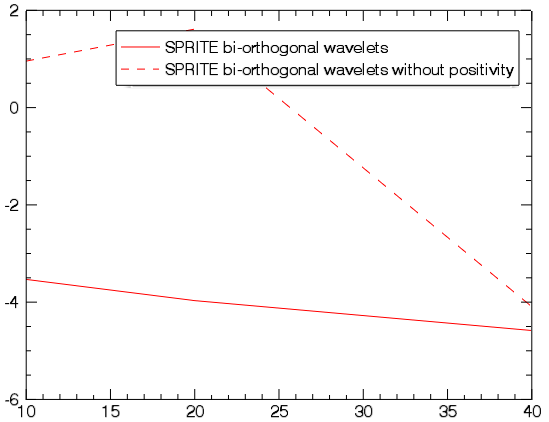}} \\
\subfigcapskip = 5pt
\subfigure[Mean absolute error on the second ellipticity parameter.]{\includegraphics[width=7cm]{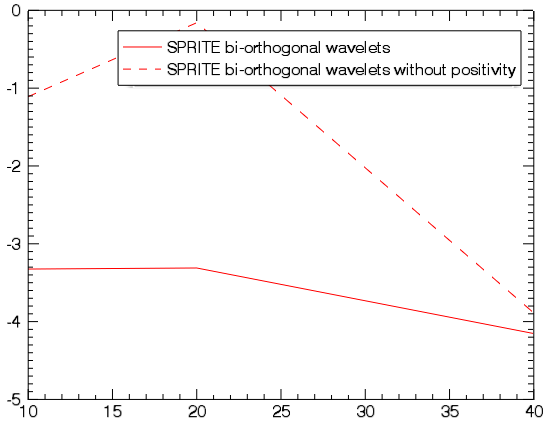}} &	
\subfigcapskip = 5pt
\subfigure[Standard deviation of the absolute error on the second ellipticity parameter.]{\includegraphics[width=7cm]{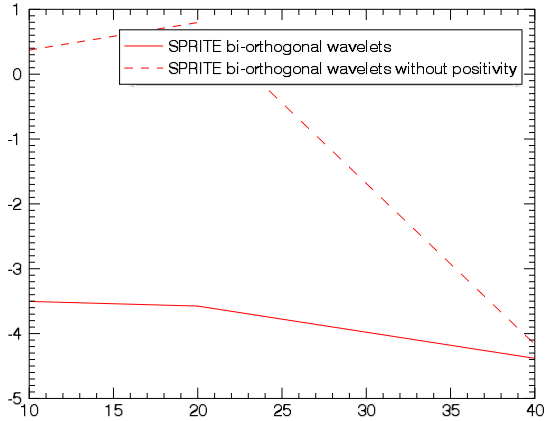}} 
\end{tabular}
\end{center}
\caption{Errors in log scale on ellipticity parameters versus the SNR. The positivity constraint significantly improves the accuracy.}

\label{num_res_2_pos}
\end{figure*}

\begin{figure*}[ht!]
\begin{center}
\includegraphics[scale=0.45]{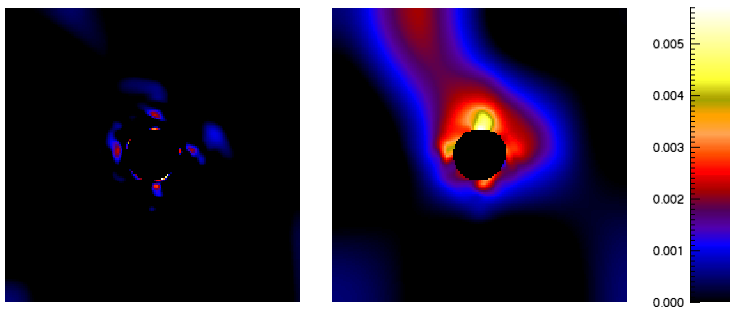}
\end{center}
\caption{Map of negative values in the PSF reconstruction map (in absolute value) at 15dB. On the left, SPRITE has a positivity constraint; on the right, SPRITE without positivity constraint.}
\label{pos_cons_illus}
\end{figure*}

We can drop the positivity constraint by simply solving
\begin{equation}
\underset{\bm{\Delta}}\min J_1(\bm{\Delta}+\mathbf{x}^{(0)}) +\kappa\|\mathbf{w}^{(k)}\odot\bm{\lambda}\odot\bm{\Phi}\bm{\Delta}\|_1
\label{no_pos_eq}
\end{equation}
at step 6 in Algorithm \ref{sprite_ana_2}. We did a similar numerical experiment as the one presented in Section \ref{num_exp} to quantify the impact of this constraint. The comparison is given in Fig. \ref{num_res_2_pos}. One can see that the positivity constraint is actually important from the point view of the ellipticity parameters at low SNR. This result is illustrated in Fig. \ref{pos_cons_illus}. With the positivity constraint, the reconstruction is less influenced by the negative oscillations in the data, due to noise; thus it yields a better robustness. Even if the negative residual values in the final PSF are 1000 order of magnitude smaller than the peak value, it is sufficient to considerably bias the ellipticity measurements.  

\section{Reweighting}
\label{rw}
\begin{figure}[h]
\begin{center}
\includegraphics[scale=0.07]{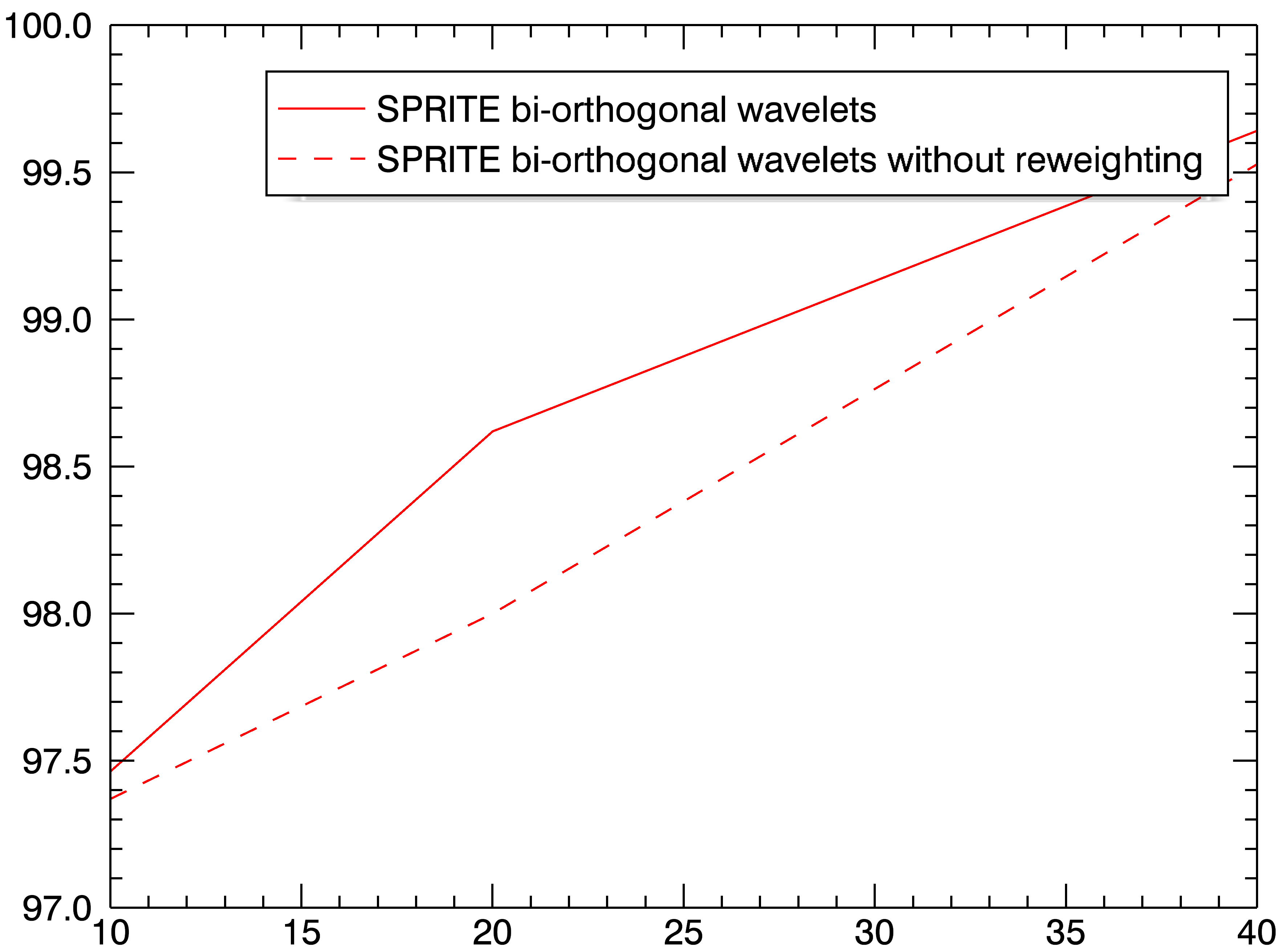}
\end{center}
\caption{Mean correlation coefficients ($\times 100$) between the SPRITE PSF reconstructions and the reference images versus the SNR; the reweighting reduces the global bias}
\label{rw_fig}
\end{figure}

As stated in Section \ref{method}, the reweighting scheme used in SPRITE is meant to mitigate the bias due to the $l_1$ norm penalty. To verify this, we basically did the same as for the positivity in the previous section. Thus, we ran Algorithm \ref{sprite_ana_2} with $K_{max} = 1$ and $K_{max} = 2$ and we compute in each case the mean correlation coefficient in Pearson sense \citep{corr_coeff} between the reference images and the reconstructions for different SNR. The result is given in Fig. \ref{rw_fig}. As expected, the reweighting improves the correlation and consequently, reduces the global bias on the reconstruction.

\bibliographystyle{aa}
\bibliography{article}

\end{document}